\documentclass[runningheads,a4paper]{llncs}

\usepackage{amsfonts}

\usepackage{pifont}

\usepackage{amssymb}
\usepackage{graphicx}

\usepackage{url}
\urldef{\mailsa}\path|pablo.lopez@ehu.es|

\newcommand{\keywords}[1]{\par\addvspace\baselineskip
\noindent\keywordname\enspace\ignorespaces#1}

\begin{document}

\mainmatter  % start of an individual contribution

\title{Are SNOMED CT Browsers Ready for Institutions? Introducing MySNOM}% Careful! Need to change it also in \titlerunning and \toctitle

\titlerunning{Are SNOMED CT Browsers Ready for Institutions? Introducing MySNOM}

\author{Pablo L\'opez-Garc\'ia} % Careful! Need to change it also in \authorrunning and \tocauthor
\authorrunning{Pablo L\'opez-Garc\'ia}

\institute{Faculty of Computer Science, University of the Basque Country\\
Paseo Manuel de Lardiz\'abal 1, Donostia-San Sebasti\'an, Spain\\
\mailsa\\
\url{http://bdi.si.ehu.es}}


\toctitle{Are SNOMED CT Browsers Ready for Institutions? Introducing MySNOM}
\tocauthor{Pablo L\'opez-Garc\'ia}
\maketitle

\begin{abstract}

SNOMED Clinical Terms (SNOMED CT) is one of the most widespread ontologies in the life sciences, with more than 300,000 concepts and relationships, but is distributed with no associated software tools. In this paper we present MySNOM, a web-based SNOMED CT browser. \mbox{MySNOM} allows organizations to browse their own distribution of SNOMED CT under a controlled environment, focuses on navigating using the structure of SNOMED CT, and has diagramming capabilities.

\end{abstract}

\section{Introduction}

The Systematized Nomenclature of Medicine - Clinical Terms (SNOMED CT) has become a reference terminology  \cite{Freitas2009} and an active research topic in both the healthcare and the Semantic Web communities \cite{Spackman2004} \cite{Heja2008}. An effective way to advance research in ontologies for the life sciences in general, and in SNOMED CT in particular, is by providing appropriate software to researchers and practitioners. In particular, visualization and browsing tools can provide a better understanding of the structure and complexity of a biomedical terminology. There are lists of existing browsers, such as the ones provided by The National Library of Medicine of the United States\footnote{\url{www.nlm.nih.gov/research/umls/Snomed/snomed_browsers.html}} and the United Kingdom's National Health Service\footnote{\url{www.connectingforhealth.nhs.uk/systemsandservices/data/snomed/browser}}, in order to provide researchers and practitioners with a reference. These lists, however, are not comprehensive, not up to date, and have not been fully ranked or evaluated. These lists can also be misleading for practitioners: as an example, Prot\'eg\'e\footnote{\url{protege.stanford.edu}} is listed as a SNOMED CT browser. In 2008, Rogers \cite{Rogers2008} performed a study of existing SNOMED CT browsers and developed a general catalog of desirable browsing features.

In this paper we present a SNOMED CT web-based browser called MySNOM, aiming at showing the structure of SNOMED, with support for rich diagrams.

\section{Architecture Overview}

\begin{figure}
\centering
\includegraphics[width=8cm]{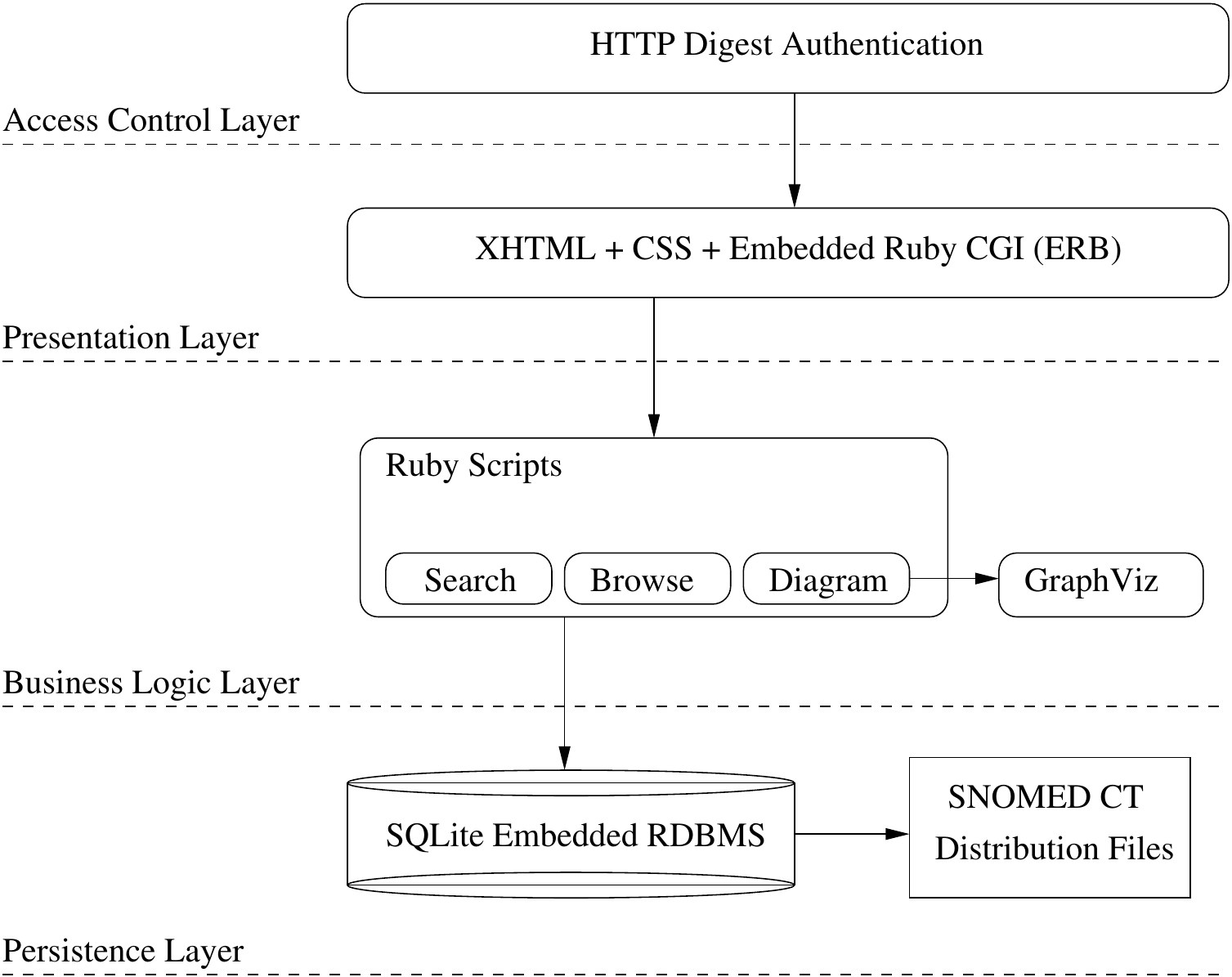}
\caption{MySNOM architecture.}
\label{fig:architecture}
\end{figure}

MySNOM is a web application following a layered approach to decouple orthogonal functionalities such as access control, graphical interface, business logic and persistence. The architecture is shown in Figure \ref{fig:architecture}.

MySNOM is hosted in a CGI-enabled web server and accessed using a web browser. It relies on the lightweight and standard HTTP Digest protocol \cite{Digest} for authentication of users, and SQLite\footnote{\url{www.sqlite.org}} to provide fast access to the terminology.

\section{User Interface and Diagramming Support}

\begin{figure}
\centering
\includegraphics[width=8cm]{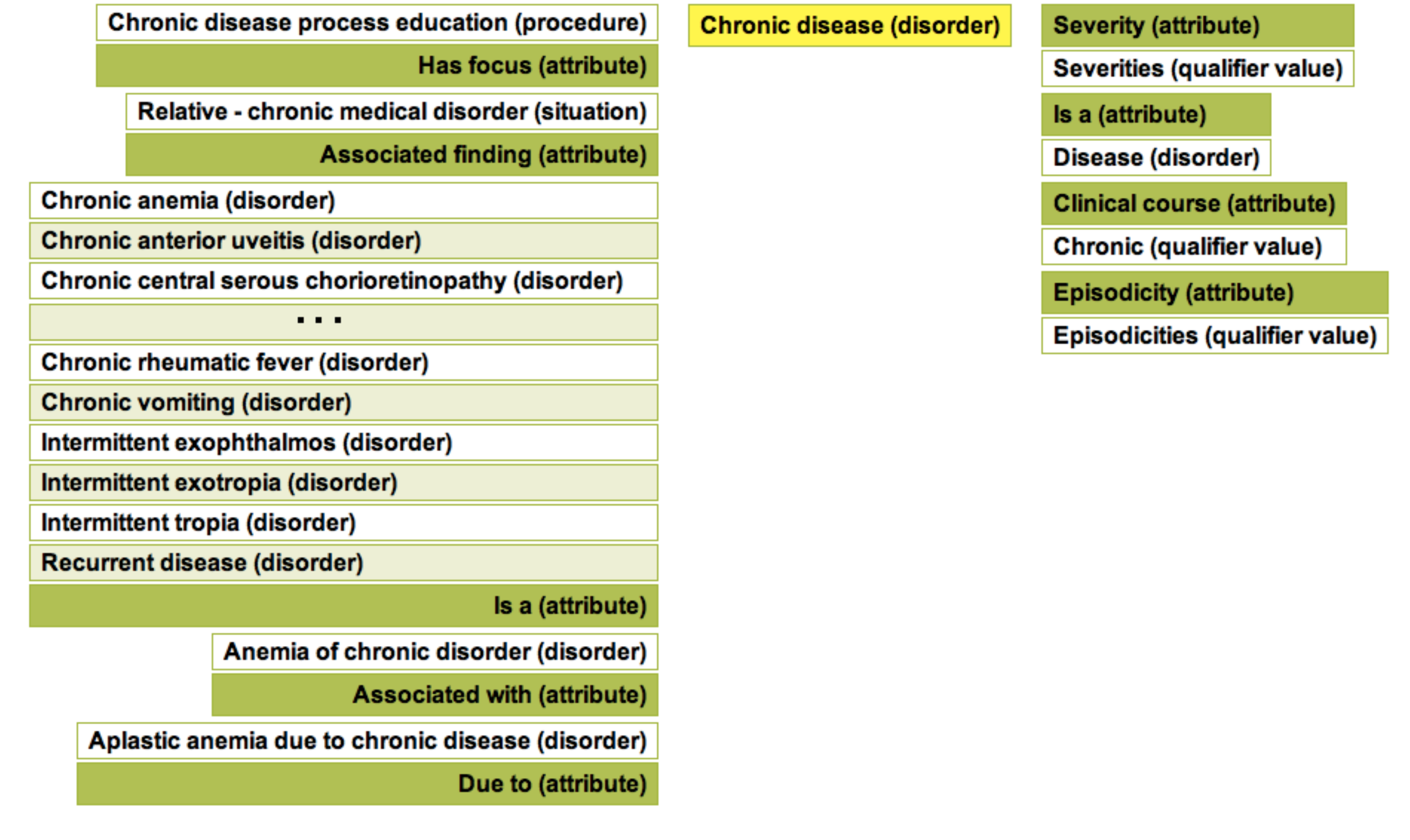}
\caption{MySNOM browsing the term `chronic disease'.}
\label{fig:browse}
\end{figure}

MySNOM uses a clean interface to explore all the relationships related to a given concept, including reverse relationships. It places the current concept in the middle of the screen, with referring concepts on the left and referred concepts on the right. This structure follows the extended left-to-right direction of flow and shows the natural directed graph structure of SNOMED CT. Every concept and relationship is clickable. Each clicked concept or relationship leads to a new screen, becoming the current term. The current concept is shown only once on screen. Users can hover their mouse over any given concept to further inspect the concept (e.g. see the SNOMED CT code). A screenshot of MySNOM browsing the term `chronic disease' is shown in Figure \ref{fig:browse}.

\begin{figure}
\centering
\includegraphics[width=10cm]{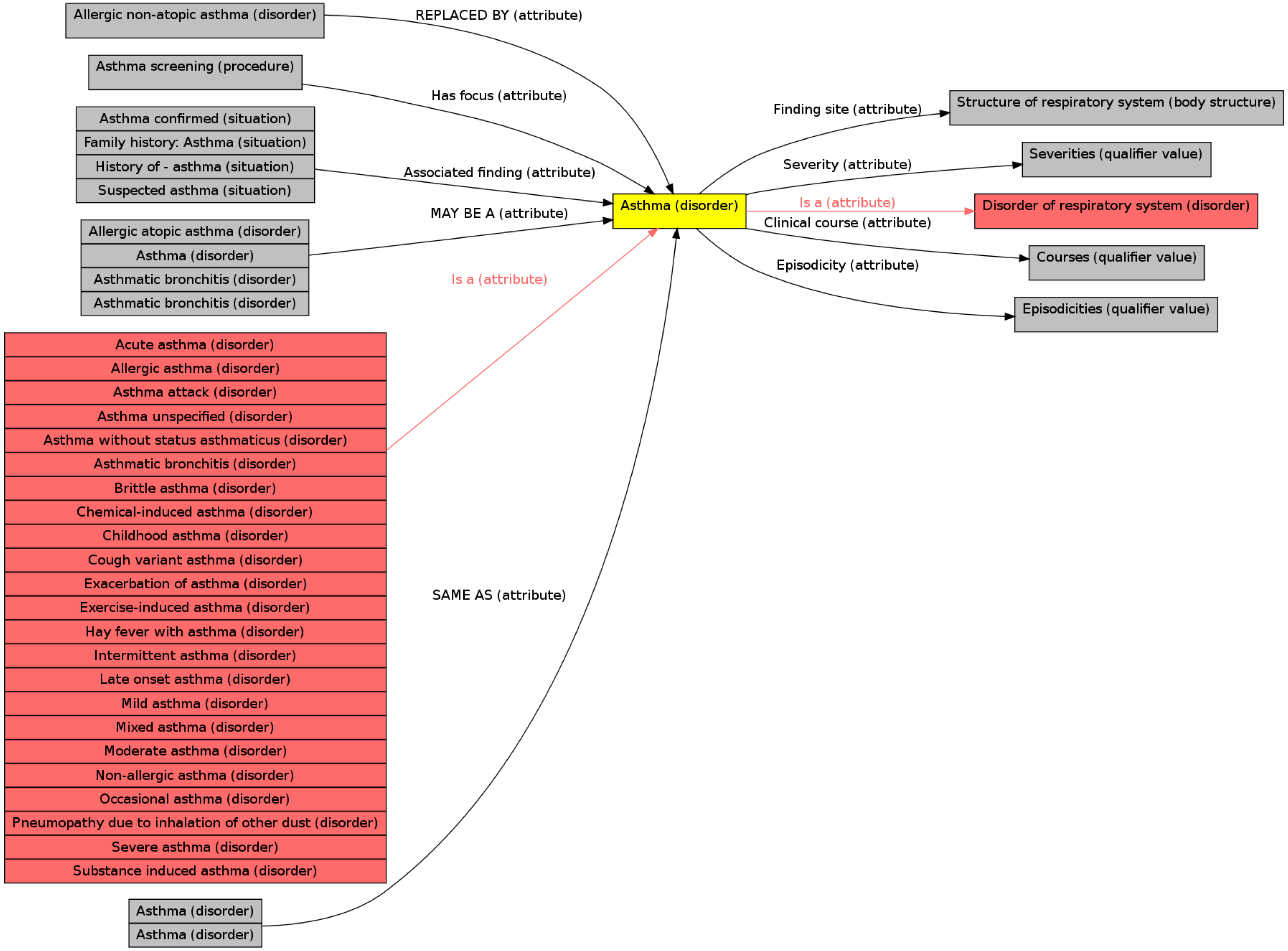}
\caption{Diagram generated by MySNOM when browsing the concept `asthma'.}
\label{fig:asthma}
\end{figure}

MySNOM supports the generation of diagrams that might clarify the context of a given concept or serve as a graphic resource to be embedded in SNOMED CT-related documents. It uses different colors to emphasize several concepts: the main concept is marked in yellow and hierarchy relationships are colored in red. The diagramming module creates an intermediate representation of the browsed concept using the DOT language \cite{Dot}. The DOT language is a plain text description that can be read by humans and processed by visualization tools. MySNOM uses Graphviz \cite{Graphviz}, an Open Source, fast, graph generation software, to generate the final diagram. Generated diagrams reflect the ontological nature of SNOMED CT, as a directed graph, showing all concepts and relationships of a given concept. A sample diagram is shown in Figure \ref{fig:asthma}.

\section{Conclusion and Future Work}

We have presented an access-controlled, web-based, lightweight browser in order to provide structural navigation and concise diagrams for SNOMED CT, using an authorized standard distribution of SNOMED CT. The browser can be viewed and tested at \cite{MySNOM} and freely obtained by contacting the author. As future work we plan to improve the prototype and add modularization support. Modularization support will let a user store and export selected portions of SNOMED CT by using ontology modularization techniques \cite{Stuckenschmidt2009}. A tool with these characteristics will allow a deep research of SNOMED CT's structure and capabilities.

\subsubsection*{Acknowledgments.} Thanks to Arantza Illarramendi, Jes\'us Berm\'udez and Idoia Berges for their support, inquiries and feedback about MySNOM. Thanks to Stefan Schulz and Martin Boeker at IMBI Freiburg for their feedback and suggestions. This work is supported by grant TIN2007-68091-C02-01 from \mbox{Ministerio} de Ciencia e Innovaci\'on (MICINN) of the Spanish Government.

\end{document}